\documentclass[journal]{IEEEtran}
\IEEEoverridecommandlockouts

\usepackage{cite}
\usepackage{amsmath,amssymb,amsfonts}
\usepackage{algorithmic}
\usepackage{graphicx}
\usepackage{textcomp}
\usepackage{xcolor}
\usepackage[colorlinks,linkcolor=red,anchorcolor=red,citecolor=blue]{hyperref}
\usepackage[lined,boxed,ruled]{algorithm2e}

\usepackage{tikz}
\usetikzlibrary{shapes.geometric, arrows.meta, positioning, fit, backgrounds, calc}

\graphicspath{{./figs/}}

\def\BibTeX{{\rm B\kern-.05em{\sc i\kern-.025em b}\kern-.08em
    T\kern-.1667em\lower.7ex\hbox{E}\kern-.125emX}}
\begin{document}

\title{MMAO: A Metabolic Multi-Agent Optimizer with Endogenous Resource Allocation for Continuous and Discrete Optimization}

\author{Jinliang Xu$^{*}$ and Liping Ma%
\thanks{Jinliang Xu received the bachelor's and master's degrees from Beijing University of Posts and Telecommunications, Beijing, China, where he also received the Ph.D. degree in computer science in 2020. He is with the China Academy of Information and Communications Technology, Beijing, China. His current research interests include heuristic optimization, agent-based applications, and blockchain systems. His e-mail: jlxufly@gmail.com and xujinliang@caict.ac.cn.}%
\thanks{Liping Ma is with the Department of Disease Control and Prevention, The Seventh Medical Center of Chinese PLA General Hospital, Beijing, China; e-mail: lipingmaqzx@163.com.}}

\maketitle
\begin{abstract}
Traditional meta-heuristics often rely on fixed population sizes, manually chosen search scales, and externally attached parameter-control modules. This paper presents the \textit{Metabolic Multi-Agent Optimizer} (MMAO), a cross-domain optimization framework in which adaptation is derived endogenously from a private-public metabolic resource loop. Each agent carries internal energy, a continuous role state, motion or structural memory, and local search history, while the population shares a communal resource pool. Fitness improvements are converted into normalized metabolic gains through a robust progress scale and a recent success statistic; the same closed loop then regulates sensing intensity, search amplitude, role drift, branching, pruning, respawning, and elite reinvestment. In the continuous setting, MMAO uses energy-regulated symmetric zero-order probing and role-interpolated motion. In the discrete setting, the same control law is instantiated through structural sensing, local route improvement, guided perturbation, and energy-weighted edge reuse. The paper combines an implementation-faithful formulation with a reproducible experimental study on a CEC2017 subset (10D/30D, 20 seeds) and five TSPLIB instances (100 discrete runs in total). The current evidence supports MMAO primarily as a parameter-light, self-calibrating optimization framework whose main validated originality lies in metabolically endogenous resource allocation across heterogeneous search behaviors, rather than as a universally superior optimizer.
\end{abstract}


\section{Introduction}
\label{sec:introduction}
\IEEEPARstart{G}{lobal} optimization of non-convex, high-dimensional, and multimodal objective functions remains a cornerstone challenge in computational intelligence and engineering applications. The fundamental dilemma in designing effective meta-heuristics lies in the dynamic trade-off between \textit{exploration} (searching the global space to avoid local optima) and \textit{exploitation} (refining solutions locally for high precision) \cite{jiang2023knowledge, shao2025knowledge}. 

\begin{figure}[!ht]
\centering
\includegraphics[width=\columnwidth]{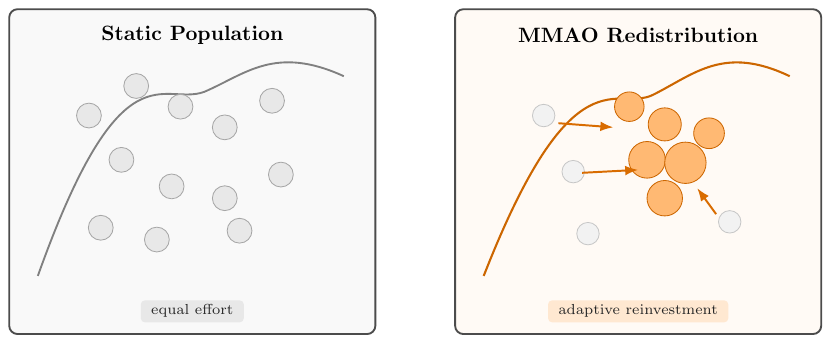}
\caption{Conceptual comparison of computational resource allocation strategies. (a) Traditional static population models (e.g., PSO, DE). (b) The proposed Metabolic Multi-Agent Optimizer (MMAO).}
\label{fig:concept_comparison}
\end{figure}

Traditional population-based algorithms, such as Particle Swarm Optimization (PSO) \cite{kennedy1995particle} and Differential Evolution (DE), typically maintain a \textit{static population size} throughout the optimization process \cite{doerr2025speeding, antipov2024already, li2025scalable, bian2025archive, bian2025stochastic}. This rigid structure treats computational resources (i.e., function evaluations) as a fixed allocation averaged across all agents. However, from the perspective of \textit{computational economics}, this approach is inherently inefficient: computational budgets are wasted on agents traversing unpromising regions (low fitness gradients), while agents in high-potential regions lack sufficient reinforcements to perform dense searching (Fig. \ref{fig:concept_comparison}). Although modern variants like L-SHADE\cite{Tanabe2014} or CMA-ES\cite{Hansen2001} introduce adaptive parameter control strategies, they rarely address the \textit{structural adaptation} of the optimizer itself \cite{cho2025configx, zhang2025laos, song2024reinforced}. More broadly, the parameter-control literature has long shown that adaptation is central to evolutionary efficiency, but in most algorithms it remains an explicit supervisory layer rather than an endogenous resource economy \cite{eiben1999parameter, karafotias2015parameter}. Specifically, there is a lack of mechanisms that autonomously regulate the population lifespan and density based on the real-time ``return on investment'' (ROI) of the search process \cite{dong2025effective, li2022distributed, liu2022cooperative, chu2024competitive, han2022multitask}. Furthermore, most existing meta-heuristics are strictly bound to either continuous or discrete domains, requiring significant re-engineering of algebraic operators when crossing the gap between functional landscapes and combinatorial structures \cite{elorza2024transforming, shao2025knowledge}.

To address this structural inefficiency, we move from static swarm intelligence toward an \textit{energy-driven multi-agent system} view \cite{liu2025less, mai2025constructive, sendra2025evolution, han2025robust, chen2024multi, zhang2024virtual}. Inspired by the morphological plasticity of plant root systems---which dynamically allocate biomass to resource-rich soil patches while pruning roots in barren areas---we introduce a metabolic framework for optimization in which every search behavior must be financed, rewarded, and recycled through a single resource loop. Each agent carries an internal energy reserve \cite{pedreschi2025human, leuzzi2025lifelong}; successful actions replenish that reserve, while maintenance, sensing, and failed search consume it. At the population level, part of the gain is reinvested into a communal pool, enabling the algorithm to expand, contract, refine, and respawn without an externally prescribed schedule.

Based on these principles, this paper presents the \textit{Metabolic Multi-Agent Optimizer} (MMAO). The current version of MMAO is not a hard-coded two-species heuristic. Instead, heterogeneous behavior emerges from a continuous role state, normalized progress feedback, and a unified communal budget. In the continuous setting, MMAO combines metabolically regulated symmetric probing with role-interpolated motion. In the discrete setting, it maps the same control logic to structural sensing over permutations, local route improvement, guided perturbation, and energy-weighted edge reuse. The resulting method is intended to be \textit{parameter-light} rather than naively parameter-free: key search scales and lifecycle decisions are generated endogenously from the metabolic loop rather than fixed by problem-specific schedules.

The paper should therefore be read as a framework-establishing contribution rather than as a final benchmark verdict. Its main value is to define the controller clearly enough that later theoretical, empirical, and domain-derivation studies can test the same object rather than loosely related variants.

The main contributions of this paper are summarized as follows:

\begin{enumerate}
    \item \textbf{A closed-loop metabolic controller:} We formulate MMAO so that normalized reward, success feedback, scale adaptation, distribution shaping, lifecycle control, and search-budget reallocation all arise from the same private-public resource accounting loop \cite{doerr2025speeding, liu2025less, dong2025effective, li2022distributed, liu2022cooperative, antipov2024already, chu2024competitive, han2022multitask, leuzzi2025lifelong}.
    \item \textbf{Continuous heterogeneous behavior without hard species:} We replace rigid role switching with a continuous role state that drives sensing density, motion composition, and reinvestment preference, treating role drift, scale adaptation, and success feedback as consequences of the same metabolic controller rather than as independent add-ons.
    \item \textbf{A unified cross-domain formulation:} We map the same MMAO logic to continuous black-box search and to discrete combinatorial optimization, using symmetric zero-order probing in Euclidean spaces and structural sensing with energy-weighted edge memory in TSP-like domains \cite{wang2023multiobjective, xu2023cumulative, fu2023hierarchical, zhao2024pega, verdu2025scaling, ye2023deepaco}.
    \item \textbf{Mechanism-level theory instead of over-claimed convergence:} We analyze bounded energy, communal reinvestment, normalized gains, and endogenous turnover as the core stabilizing mechanisms of MMAO, rather than claiming a complete theorem for the full adaptive nonlinear system \cite{doerr2023understanding, dang2025dominance, zheng2023runtime, vermetten2024large, cenikj2025comparing, lehre2024concentration, omeradzic2024self, prager2024exploratory, stripinis2024benchmarking, raponi2023optimizing, zheng2024approximation}.
    \item \textbf{An implementation-faithful experimental study:} We reorganize the paper around the current MMAO implementation and provide a reproducible experimental section covering CEC-style continuous tests, five TSPLIB benchmarks, formal nonparametric comparisons, and mechanism-level diagnostic traces, while keeping the structure extensible for larger future benchmark campaigns.
\end{enumerate}

The remainder of this paper is organized as follows: Section II reviews related work in adaptive meta-heuristics and plant-inspired optimization. Section III reformulates the current MMAO method in an implementation-faithful way, including its continuous and discrete instantiations. Section IV presents a mechanism-level theoretical analysis. Section V reports the current experimental study and explains how the benchmark pipeline can be expanded further. Section VI discusses the methodological position, parameter-light ambition, and current limitations of MMAO. Section VII concludes the paper.

\section{Related Work}
The related work covers adaptive evolutionary algorithms and plant-inspired meta-heuristics, with emphasis on the differences between MMAO and existing root-based algorithms.

\subsection{Adaptive Evolutionary Algorithms}
Global optimization has long been dominated by classic methods such as Genetic Algorithms (GA) \cite{holland1975adaptation}, Simulated Annealing (SA) \cite{kirkpatrick1983optimization}, and Particle Swarm Optimization (PSO) \cite{kennedy1995particle}. While these algorithms established the foundation of stochastic search, modern variants often focus on parameter adaptation to navigate complex landscapes. 

However, a critical limitation persists: most of these algorithms, including recent bio-inspired methods like the Slime Mould Algorithm (SMA) \cite{li2020slime}, typically maintain a \textit{static population size} or use simple deterministic reduction strategies \cite{doerr2025speeding, antipov2024already, li2025scalable, bian2025archive, bian2025stochastic}. They lack an intrinsic economic mechanism to autonomously regulate the computational budget based on the real-time ``Return on Investment'' (ROI) of the search process \cite{dong2025effective, li2022distributed, liu2022cooperative, chu2024competitive, han2022multitask}.

\subsection{Plant-Inspired Meta-heuristics}
Biological root systems and plant behaviors have inspired several algorithms. Notable examples include the Artificial Plant Optimization Algorithm (APOA) \cite{cui2013artificial} and the Tree-Seed Algorithm (TSA) \cite{kiran2015tsa}, which simulate branching and seeding processes respectively.

Regarding root-specific modeling, the most relevant works are Invasive Weed Optimization (IWO) (e.g., as applied in \cite{azizipour2016optimal}), Root Mass Optimization (RMO) \cite{qi2013idea}, and the Root Growth Model \cite{zhang2012root}.

\textbf{IWO} simulates the ecological colonization of weeds but treats individuals as static entities that disperse seeds based on Gaussian distribution, lacking active agent kinematics.

\textbf{RMO and RGA}, while sharing the ``root'' metaphor, differ fundamentally from our proposed MMAO in terms of abstraction level and operational logic:
\begin{itemize}
    \item \textit{Morphological vs. Metabolic:} RMO \cite{qi2013idea} and RGA \cite{zhang2012root} primarily mimic the \textbf{geometric morphology} of roots (e.g., branching angles, density constraints) to occupy the search space. In contrast, MMAO abstracts the \textbf{metabolic economics} of roots, focusing on how energy (computational budget) is acquired, stored, and dissipated. 
    \item \textit{Static vs. Dynamic Agents:} In prior works like TSA \cite{kiran2015tsa} or RMO \cite{qi2013idea}, the branching/propagation behavior is typically governed by fixed probabilistic rules. In MMAO, the behavior is emergent: agents autonomously decide to branch, remain dormant, or die based on their internal energy state, leading to self-regulated turnover and density adaptation.
    \item \textit{Guidance Mechanism:} Traditional approaches often rely on random walk or vector-based guidance similar to PSO. MMAO introduces an energy-regulated symmetric gradient estimation, linking the root-inspired metaphor to a more principled zero-order sensing view \cite{nesterov2017random, seung2025low, hikima2025zeroth, lei2025zeroth}.
\end{itemize}

RMO and RGA model root \textit{morphology}. MMAO models resource-constrained \textit{survival} through energy acquisition and expenditure.

\subsection{Bridging Continuous and Discrete Optimization}
A longstanding challenge in meta-heuristics is the architectural gap between continuous function optimization and discrete combinatorial search \cite{elorza2024transforming, shao2025knowledge}. Conventional approaches often rely on specific mapping techniques, such as \textit{random key} encodings or the redefinition of algebraic operators into discrete swap sequences. However, these methods frequently sacrifice the intrinsic search dynamics of the original algorithm. Unlike these domain-specific adaptations, the concept of \textit{structural sensing} within a metabolic framework remains relatively unexplored. By treating the search process as a resource-constrained survival game, it is possible to maintain a unified logic across diverse topological spaces—a paradigm that MMAO aims to establish \cite{wang2023multiobjective, xu2023cumulative, fu2023hierarchical, zhao2024pega, verdu2025scaling, ye2023deepaco}.

\section{The MMAO Framework}
\begin{figure}[!h]
\centering
\includegraphics[width=\columnwidth]{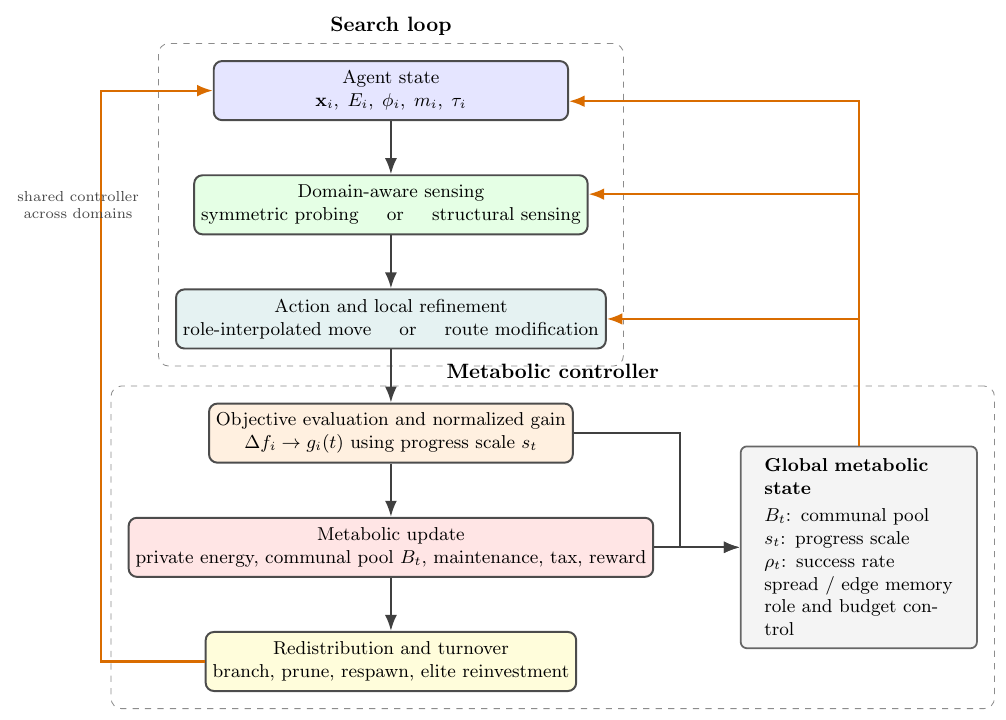}
\caption{The schematic framework of MMAO. It illustrates the coupled dynamics between the micro-level perception-action loop and the macro-level metabolic lifecycle.}
\label{fig:architecture}
\end{figure}

We describe the current MMAO implementation as a closed-loop control system in which metabolically endogenous resources finance and regulate search behavior. The overall architecture is illustrated in Fig. \ref{fig:architecture}. The method couples three layers: (1) agent-level state and sensing, (2) private-public resource accounting, and (3) population-level redistribution and turnover.

\subsection{System Modeling and Problem Definition}
We consider a global minimization problem
\begin{equation}
    \min_{\mathbf{x} \in \Omega} f(\mathbf{x}),
\end{equation}
where $\Omega \subseteq \mathbb{R}^D$ in the continuous case, or where $\mathbf{x}$ denotes a discrete combinatorial structure such as a permutation in the TSP case. MMAO does not rely on one domain-specific algebraic operator. Instead, it treats each candidate solution as the local embodiment of a resource-constrained agent. The mapping between the root-inspired metaphor and the computational formulation is summarized in Table \ref{tab:mapping}.

\begin{table}[h]
\caption{Mapping from Biological Roots to MMAO}
\label{tab:mapping}
\centering
\begin{tabular}{ll}
\hline
\textbf{Biological Concept} & \textbf{Computational Equivalent} \\
\hline
Root Tip & Autonomous Search Agent \\
Soil Environment & Objective Function Landscape $f(\mathbf{x})$ \\
Nutrients/Water & Negative Fitness Value $-f(\mathbf{x})$ \\
Photosynthesis & Fitness-based Reward Mechanism \\
Metabolic Decay & Regularization / Maintenance Cost \\
Biomass & Internal Energy State ($E$) \\
\hline
\end{tabular}
\end{table}

\subsection{Agent State and Global Metabolic State}
Let $\mathcal{P}_t = \{A_1, A_2, \dots, A_{N_t}\}$ be the active population at iteration $t$. In the current MMAO implementation, each agent $A_i$ is defined by
\begin{equation}
    S_i(t) = \langle \mathbf{x}_i, \mathbf{v}_i, E_i, \phi_i, \tau_i, m_i \rangle,
\end{equation}
where $\mathbf{x}_i$ is the current position or structure, $\mathbf{v}_i$ is a momentum-like displacement memory in continuous search, $E_i \ge 0$ is the private energy reserve, $\phi_i \in [0,1]$ is a continuous role state, $\tau_i$ is the age, and $m_i$ denotes local memory such as the personal best solution or best route. The role variable does not encode a hard agent class. Instead, it continuously interpolates between exploration-dominant and exploitation-dominant behavior. Values $\phi_i \approx 0$ correspond to broader probing and stronger perturbation, while values $\phi_i \approx 1$ correspond to denser refinement and stronger memory-guided contraction.

The population also shares a \textit{global metabolic state}
\begin{equation}
    G_t = \langle B_t, s_t, \rho_t \rangle,
\end{equation}
where $B_t$ is a unified communal resource pool, $s_t$ is a robust scale estimate extracted from recent positive improvements, and $\rho_t$ is the recent success rate. These variables summarize the current metabolic condition of the system. In the implementation, they are recomputed online and reused to drive sensing density, move amplitude, branching intensity, acceptance leniency, and respawn bias.

\subsection{Metabolic Closure and Normalized Reward}
The key design principle of MMAO is that adaptive behaviors must be generated by the metabolic loop itself rather than inserted as disconnected controllers. For a minimization problem, define the raw improvement
\begin{equation}
    \Delta f_i(t) = f(\mathbf{x}_i(t)) - f(\mathbf{x}_i(t+1)).
\end{equation}
Instead of reacting directly to raw gains, MMAO normalizes them through the endogenous progress scale:
\begin{equation}
    g_i(t) = \frac{\max(0,\Delta f_i(t))}{\varepsilon + s_t}.
\end{equation}
The quantity $g_i(t)$ is then mapped to a bounded reward pathway, typically through a clipped logarithmic transformation, while the maintenance charge $C_i(t)$ is derived from role intensity, crowding, success scarcity, and stagnation. In this sense, MMAO embeds parameter control and operator allocation into one metabolic controller rather than into a separate scheduling module \cite{karafotias2015parameter, li2014adaptivebandit}. The private energy update becomes
\begin{equation} \label{eq:energy}
    E_i(t+1) = \Pi_{[0,E_{\max}]}\left(E_i(t) + (1-\alpha_i)R_i(t) - C_i(t)\right),
\end{equation}
where $\alpha_i \in [0,1]$ is the communal contribution ratio and $\Pi$ denotes clipping to the admissible interval. The communal pool is updated by
\begin{equation}
    B_{t+1} = \max\left\{0, B_t + \sum_i \left(C_i(t) + \alpha_i R_i(t)\right) - U_t \right\},
\end{equation}
where $U_t$ is the amount reinvested into branching, respawning, guided reconstruction, or elite reinforcement. This is the central closure property of MMAO: successful search not only improves the objective value, but also generates the resources that determine future allocation of search effort.

\subsection{Endogenous Scale and Distribution Adaptation}
The variables $s_t$ and $\rho_t$ induce two self-calibration effects that are explicitly reflected in the current implementation.
\begin{itemize}
    \item \textbf{Scale adaptation:} When recent progress becomes small, the normalization scale becomes tighter, making weaker improvements metabolically visible. At the same time, a low success rate increases exploratory pressure. As a consequence, probing radii, perturbation intensity, local search span, and reinvestment aggressiveness expand or contract without requiring a hand-crafted schedule.
    \item \textbf{Distribution adaptation:} MMAO does not fit a separate global density model. Instead, the surviving population defines an energy-weighted empirical distribution. In continuous spaces this appears as adaptive spatial spread around promising regions. In TSP-like spaces it appears as energy-weighted edge memory extracted from metabolically successful routes.
\end{itemize}

\subsection{Energy-Regulated Sensing}
Before generating an action, each agent allocates part of its current budget to sensing. The sensing intensity is determined by relative metabolic condition:
\begin{equation}
    K_i(t) = K_{\min} + \left\lfloor (K_{\max}-K_{\min}) \Psi(E_i, \phi_i, \rho_t, \tau_i) \right\rfloor,
\end{equation}
where $\Psi(\cdot)\in[0,1]$ is a smooth metabolic drive function. Agents with higher surplus can afford denser local measurement, while low-success regimes increase broader exploratory sensing and larger structural trials.

For continuous optimization, sensing is implemented by symmetric zero-order probing \cite{seung2025low, hikima2025zeroth, lei2025zeroth}:
\begin{equation}\label{eq:grad}
    \hat{\mathbf{g}}_k = \frac{f(\mathbf{x}_i + r_i \mathbf{d}_k) - f(\mathbf{x}_i - r_i \mathbf{d}_k)}{2r_i} \cdot \mathbf{d}_k,
\end{equation}
where $\mathbf{d}_k$ is a random unit vector and $r_i$ is an adaptive sensing radius derived from current spread, success scarcity, role state, and energy surplus. The implementation uses multiple biased random directions around the agent and aggregates them into a normalized directional estimate. Thus both sample count and probing scale emerge from the same budget logic.

\subsection{Continuous Role Dynamics and Motion Generation}
MMAO does not hard-code an exploratory class and an exploitative class. Instead, the role state evolves continuously:
\begin{equation}
    \phi_i(t+1) = \Pi_{[0,1]}\left(\phi_i(t) + \beta \cdot \Xi(g_i(t), \rho_t, E_i(t), \tau_i)\right),
\end{equation}
where $\Xi(\cdot)$ is a signed metabolic feedback term. Agents that convert local opportunities into efficient gain drift toward stronger exploitation, whereas unsuccessful, under-supplied, or stagnant agents drift back toward exploration. In the implementation, this update also depends on normalized reserve level and stagnation counters.

The movement itself is generated by role interpolation rather than role switching. In the continuous case, an abstract update can be written as
\begin{equation}
\begin{aligned}
    \Delta \mathbf{x}_i ={}& \omega_i \mathbf{v}_i - \lambda_i(\phi_i)\hat{\mathbf{g}}_i \\
    &+ \mu_i(\phi_i)(\mathbf{x}^{best}_i-\mathbf{x}_i) + \nu_i(\phi_i)(\mathbf{x}^{g}-\mathbf{x}_i) \\
    &+ \sigma_i(\phi_i)\boldsymbol{\xi}_i,
\end{aligned}
\end{equation}
where $\mathbf{x}^{best}_i$ is the personal best, $\mathbf{x}^{g}$ is the global best, and $\boldsymbol{\xi}_i$ is exploratory noise. Small $\phi_i$ emphasizes momentum and broad perturbation, whereas large $\phi_i$ emphasizes local refinement and memory-guided contraction. The current implementation additionally allows a bounded uphill acceptance policy, so unsuccessful but metabolically viable agents can occasionally preserve exploratory moves when the normalized loss is small.

\begin{algorithm}[!t]
\caption{Metabolic Multi-Agent Optimizer (MMAO)}
\label{alg:MMAO}
\begin{algorithmic}[1]
\REQUIRE 
    Objective function $f(\mathbf{x})$ or a discrete structural cost, search domain $\Omega$.
ENSURE Global best solution $\mathbf{x}^*$.

\STATE \textbf{Initialize:} 
\STATE Generate an initial population $\mathcal{P}_0$ with agent energies $E_i$, role states $\phi_i$, and communal pool $B_0$.
\STATE Set global best $\mathbf{x}^* \leftarrow \arg\min_{\mathbf{x} \in \mathcal{P}_0} f(\mathbf{x})$.
\STATE $t \leftarrow 0$.

\WHILE{Termination criterion not met}
    \STATE Estimate the progress scale $s_t$, success rate $\rho_t$, and energy-weighted population distribution.
    \FOR{each active agent $A_i \in \mathcal{P}_t$}
        \STATE Derive sensing budget and sensing scale from the current metabolic state.
        \STATE Perform continuous or structural sensing to generate candidate moves.
        \STATE Construct the accepted move by interpolating exploratory and exploitative actions through $\phi_i$.
        \STATE Evaluate the candidate and compute normalized gain $g_i$.
        \STATE Update private energy $E_i$, communal pool $B_t$, age $\tau_i$, and role state $\phi_i$.
        \IF{$f(\mathbf{x}_i(t+1)) < f(\mathbf{x}^*)$}
            \STATE $\mathbf{x}^* \leftarrow \mathbf{x}_i(t+1)$.
        \ENDIF
    \ENDFOR
    
    \STATE Prune agents whose private reserve drops below the metabolically admissible floor.
    \STATE If an agent exhibits sustained surplus and recent success, allocate shared resources for local branching.
    \STATE If the population becomes too sparse, respawn exploratory agents using the communal pool and the current search distribution.
    \STATE $t \leftarrow t + 1$.
\ENDWHILE
\RETURN $\mathbf{x}^*$.
\end{algorithmic}
\end{algorithm}

\subsection{Population Lifecycle from a Unified Resource Pool}
The population size $N_t$ remains adaptive, but its evolution is mediated by one communal budget rather than by several disconnected thresholds.
\begin{itemize}
    \item \textbf{Pruning:} Agents that fail to maintain a viable private reserve are removed, and part of their residual energy is recycled into $B_t$.
    \item \textbf{Branching:} Agents with sustained normalized gains and positive surplus may draw from $B_t$ to seed offspring near metabolically successful regions. Offspring inherit neither a fixed role nor a fixed search scale; both are initialized from the same local metabolic context.
    \item \textbf{Respawn:} If the active population becomes too sparse, $B_t$ subsidizes new exploratory agents. Respawning may be global, elite-biased, or distribution-guided depending on current success and population structure.
    \item \textbf{Elite reinvestment:} When the communal pool is sufficiently rich, part of it may be reinvested into focused refinement around the best region or into guided route reconstruction. This is not a separate local-search module; it is funded by the same communal loop and therefore remains metabolically endogenous.
\end{itemize}
This unified resource pool is central to the parameter-light claim: MMAO avoids manually scheduled population reduction, separate success-control layers, and domain-specific reallocation logic to decide where computational effort should go.

\subsection{Computational Complexity Analysis}
The per-iteration complexity of MMAO depends on the number of active agents and on the domain-specific sensing budget. In the continuous case, let $\bar{N}$ be the average population size and let $\bar{K}$ be the average number of symmetric probe directions. The dominant cost is $O(\bar{N}\bar{K}D)$ plus additional candidate evaluations for bounded local refinement. In the TSP case, the dominant work comes from route perturbation, candidate reconstruction, and bounded 2-opt passes; if $n$ is the number of cities and $\bar{B}$ is the effective structural budget per agent, then the dominant cost is approximately $O(\bar{N}\bar{B})$ per iteration, with constants determined by local move evaluation.

The key point is not that MMAO is asymptotically cheaper than classical methods, but that the evaluation budget is redistributed dynamically. The pruning mechanism suppresses persistent low-return agents, while branching and reinvestment amplify high-return regions. Hence the effective evaluation economy is adaptive even when the per-agent sensing operator is more expensive than a simple random perturbation.

\subsection{Extension to Discrete Combinatorial Optimization: The TSP Case}
To demonstrate the cross-domain nature of MMAO, we instantiate the same metabolic control law for the Traveling Salesman Problem (TSP) \cite{wang2023multiobjective, xu2023cumulative, fu2023hierarchical, zhao2024pega, verdu2025scaling, ye2023deepaco, heins2025repair, wu2024reinforcens}. Unlike traditional adaptations that simply discretize continuous formulas, MMAO uses a \textit{structural sensing} mapping:

\subsubsection{Structural Position Mapping} In discrete manifolds, an agent's position $\mathbf{x}$ is no longer a coordinate in $\mathbb{R}^D$ but a topological permutation of cities, represented as a set of edges $\mathcal{E} = \{(c_1, c_2), (c_2, c_3), \dots, (c_n, c_1)\}$.

\subsubsection{Metabolic Structural Sensing} In the TSP case, the symmetric probing idea is replaced by local edge recomposition. The agent expends a metabolically determined structural budget on bounded 2-opt refinement, route kicks such as double-bridge perturbation, and guided route reconstruction. The same variables $E_i$, $\phi_i$, $s_t$, $\rho_t$, and $B_t$ control the number of trials, the locality of modifications, acceptance leniency, and the aggressiveness of perturbation.

\subsubsection{Distribution Adaptation in Edge Space} The discrete version of MMAO maintains an energy-weighted edge memory extracted from the surviving population and reinforced by elite edges. This memory is not an external probabilistic model; it is simply the current empirical distribution of metabolically successful sub-tours. High-energy edges become more likely to reappear in spawned, branched, or reinvested routes, while low-value structures disappear naturally through negative net energy. In this way, distribution adaptation in the discrete space remains a direct consequence of the same metabolic closure principle used in the continuous domain.

\section{Theoretical Analysis}
In this section, we analyze MMAO at the mechanism level. Because the current optimizer contains continuous role dynamics, communal reinvestment, guided respawn, and adaptive search distribution, a classical global-convergence theorem for the entire nonlinear process would either be overly weak or rest on unrealistic simplifications. We therefore focus on the stability properties that remain interpretable and defensible \cite{doerr2023understanding, dang2025dominance, zheng2023runtime, vermetten2024large, cenikj2025comparing, lehre2024concentration, omeradzic2024self, prager2024exploratory, stripinis2024benchmarking, raponi2023optimizing, zheng2024approximation}.

\subsection{Preliminaries and Assumptions}
Consider the minimization problem $\min_{\mathbf{x} \in \Omega} f(\mathbf{x})$. For the continuous interpretation, we make the following standard assumptions:
\begin{itemize}
    \item \textbf{A1 (Boundedness):} The search space $\Omega$ is compact, and the objective function $f(\mathbf{x})$ is bounded from below (i.e., $f(\mathbf{x}) > -\infty$).
    \item \textbf{A2 (Smoothness):} The gradient $\nabla f(\mathbf{x})$ exists and is Lipschitz continuous with constant $L$, i.e., $\|\nabla f(\mathbf{x}) - \nabla f(\mathbf{y})\| \le L \|\mathbf{x} - \mathbf{y}\|$.
\end{itemize}
Assumption A2 is used only to interpret the zero-order sensing mechanism and not to claim that MMAO becomes a standard gradient method. In the discrete case, the same theoretical message should be read structurally: the method operates on bounded search states, bounded resources, and finite local modification budgets.

\subsection{Mechanism-Level Stability}
The current MMAO is substantially more adaptive than an earlier hard-typed variant, so a simple classical convergence theorem is no longer the right theoretical target. Instead, we analyze the stability mechanisms that remain valid after the redesign.

\textbf{Definition 1 (Normalized Progress Signal):} The statistic $s_t$ is a rolling robust estimate of recent positive improvements and serves as an endogenous scale estimator for the optimizer.

\textbf{Observation 1 (Reward Normalization):} Because MMAO reacts to $g_i(t)=\max(0,\Delta f_i)/(s_t+\varepsilon)$ rather than to $\Delta f_i$ directly, the reward pathway is much less sensitive to absolute objective scale, which improves cross-problem robustness of the resource loop.

\textbf{Lemma 1 (Bounded Private Energy):} By construction, the clipping operator in Eq. \eqref{eq:energy} guarantees $0 \le E_i(t) \le E_{\max}$ for every agent and every iteration.

\textbf{Lemma 2 (Bounded Communal Budget Under Finite Reward):} If per-iteration reward is bounded, then the communal pool $B_t$ remains bounded whenever the reinvestment policy consumes a non-zero fraction of accumulated communal income.

\begin{IEEEproof}
Both private and communal resources are driven by bounded improvement signals under Assumption A1. Private energy is clipped explicitly, while the communal pool is repeatedly drained through branching subsidies, respawn, and elite reinforcement. Therefore, the system cannot accumulate unbounded resources unless the objective provides unbounded gains in finite time, which is excluded here.
\end{IEEEproof}

\textbf{Lemma 3 (Endogenous Turnover):} If an agent remains in a regime where normalized gain stays sufficiently small for long enough, then its net energy balance becomes negative and the agent is eventually pruned or recycled into the communal pool.

\begin{IEEEproof}
When local progress vanishes, the gain term in Eq. \eqref{eq:energy} collapses, whereas the maintenance term remains strictly positive because it includes survival, crowding, and stagnation pressure. Consequently, the agent cannot preserve its reserve indefinitely and must either recover through new progress or disappear from the active population.
\end{IEEEproof}

\subsection{Interpretation of the Theoretical Picture}
The previous lemmas do not prove global optimality for the full adaptive system. They support a more modest but more defensible claim: MMAO behaves as a bounded self-calibrating search process with persistent turnover and continual redistribution of computational effort. This is the appropriate theoretical level for the present paper because the implemented optimizer contains role interpolation, bounded uphill tolerance, communal reinvestment, guided respawn, and distribution feedback that are essential in practice but are not naturally captured by a classical Robbins-Monro style proof.

\section{Experiments}
This section reports the completed benchmark artifacts generated from the current MMAO code base (see GitHub repository \texttt{mmao}\footnote{\url{https://github.com/wolfbrother/mmao}} or PyPI package \texttt{mmao-opt}\footnote{\url{https://pypi.org/project/mmao-opt/}}). The goal is conservative: verify that the continuous and discrete instantiations both run stably, that the metabolic loop behaves as intended, and that the recorded traces support the resource-redistribution claim.

\subsection{Experimental Setup}
The completed run uses 20 fixed seeds for each benchmark. In the continuous domain, we evaluate MMAO on 10D and 30D CEC2017 settings for F1, F3, F4, F5, F6, F7, F9, and F10. In the discrete domain, the completed TSPLIB run covers \texttt{eil51}, \texttt{eil76}, \texttt{berlin52}, \texttt{kroA100}, and \texttt{st70}. The continuous summaries therefore aggregate 40 runs per function, namely 20 seeds across 2 dimensions, while the discrete summaries aggregate 100 runs per method, namely 5 instances times 20 seeds.

For continuous benchmarks, performance is reported as the final best objective value minus the official CEC bias. For TSP benchmarks, performance is reported by final tour length and percentage gap to the known optimum. We also log iteration-wise diagnostics including population size, total private energy, communal pool size, success rate, progress scale, and mean role state.

To make comparisons explicit, we use the same function/seed grid for all continuous methods and the same instance/seed grid for all TSP methods. Budget control is protocol-based rather than exact function-evaluation equalization across heterogeneous optimizers. On the continuous side, Random Search uses $80D$ evaluations, Hill Climb uses $60D$ evaluations, and DE-lite uses 70 generations with population size 14; MMAO and its continuous ablations share the same 110-iteration budget together with identical dimension-dependent population and sensing bounds. On the TSP side, NN+2opt, RR-2opt, and CI+2opt each use one constructive restart followed by bounded 2-opt, while MMAO and its discrete ablations share the same 60-iteration lifecycle budget and identical population bounds. This makes the comparisons fair at the protocol level, while still leaving exact cross-family function-evaluation normalization as future work. For aggregate comparisons, we report Mann-Whitney U tests, Cliff's delta, and common-language effect probabilities from the recorded run distributions.

\subsection{Continuous Results}
Table~\ref{tab:continuous-results} summarizes the completed continuous suite. The eight tested functions span ill-conditioned, highly multimodal, and hybrid-composition behavior.

\begin{table}[h]
\caption{Continuous results on the CEC2017 subset (10D/30D, 20 seeds). Lower is better.}
\label{tab:continuous-results}
\centering
\begin{tabular}{lcccc}
\hline
\textbf{Function} & \textbf{Runs} & \textbf{Mean} & \textbf{Std} & \textbf{Best} \\
\hline
F1  & 40 & 21193.2450 & 20423.0200 & 1263.1149 \\
F3  & 40 & 40.8629 & 38.1577 & 0.0548 \\
F4  & 40 & 134.5242 & 109.1068 & 15.6038 \\
F5  & 40 & 0.0007 & 0.0012 & 0.0000 \\
F6  & 40 & 127.9723 & 84.1445 & 32.2703 \\
F7  & 40 & 245.3965 & 206.6141 & 28.0000 \\
F9  & 40 & -2120.6557 & 895.1796 & -4811.2754 \\
F10 & 40 & 41834.2435 & 57763.2732 & 18.1239 \\
\hline
\end{tabular}
\end{table}

These results should be interpreted as evidence of viability rather than final superiority.

\subsection{Discrete Results}
Table~\ref{tab:tsp-results} reports the completed discrete run on five TSPLIB instances. The goal is not to compete with highly specialized TSP solvers, but to verify that the discrete MMAO controller remains functional on standard benchmarks beyond a single toy instance.

\begin{table}[h]
\caption{TSPLIB results for the discrete MMAO variant (5 instances, 20 seeds each). Lower gap is better.}
\label{tab:tsp-results}
\centering
\begin{tabular}{lcccc}
\hline
\textbf{Instance} & \textbf{Runs} & \textbf{Mean Gap (\%)} & \textbf{Std} & \textbf{Best Gap (\%)} \\
\hline
\texttt{berlin52} & 20 & 6.7025 & 2.8105 & 2.5192 \\
\texttt{eil51}    & 20 & 6.2207 & 1.5509 & 3.2864 \\
\texttt{eil76}    & 20 & 11.2546 & 2.4110 & 6.8773 \\
\texttt{kroA100}  & 20 & 10.3877 & 2.0548 & 7.1469 \\
\texttt{st70}     & 20 & 9.4667 & 2.3084 & 5.9259 \\
\hline
\end{tabular}
\end{table}

The discrete variant therefore works end-to-end on a small but non-trivial TSPLIB set. The resulting gaps are still far from state-of-the-art TSP performance, but they are strong enough to support the narrower claim pursued in this paper: the same closed-loop metabolic controller can be instantiated meaningfully in a discrete combinatorial domain.

\subsection{Mechanism-Level Diagnostics}
Terminal objective values alone do not test the core claim of MMAO. We therefore examine whether the recorded traces exhibit the closed-loop resource behavior described in Section III.

\subsubsection{Continuous Mechanism Example}
On the 2D Rastrigin sanity-check run, MMAO improves the best value from $9.9407$ at initialization to $4.6846 \times 10^{-4}$ after 220 iterations. More importantly, the trace reveals the intended metabolic cycle: after the early gain phase, the active population contracts toward a narrow band of 4--6 agents while the communal pool remains available for later reinvestment (Fig.~\ref{fig:mechanism-traces}a).

\subsubsection{Discrete Mechanism Example}
\begin{figure*}[!t]
\centering
\includegraphics[width=0.48\textwidth]{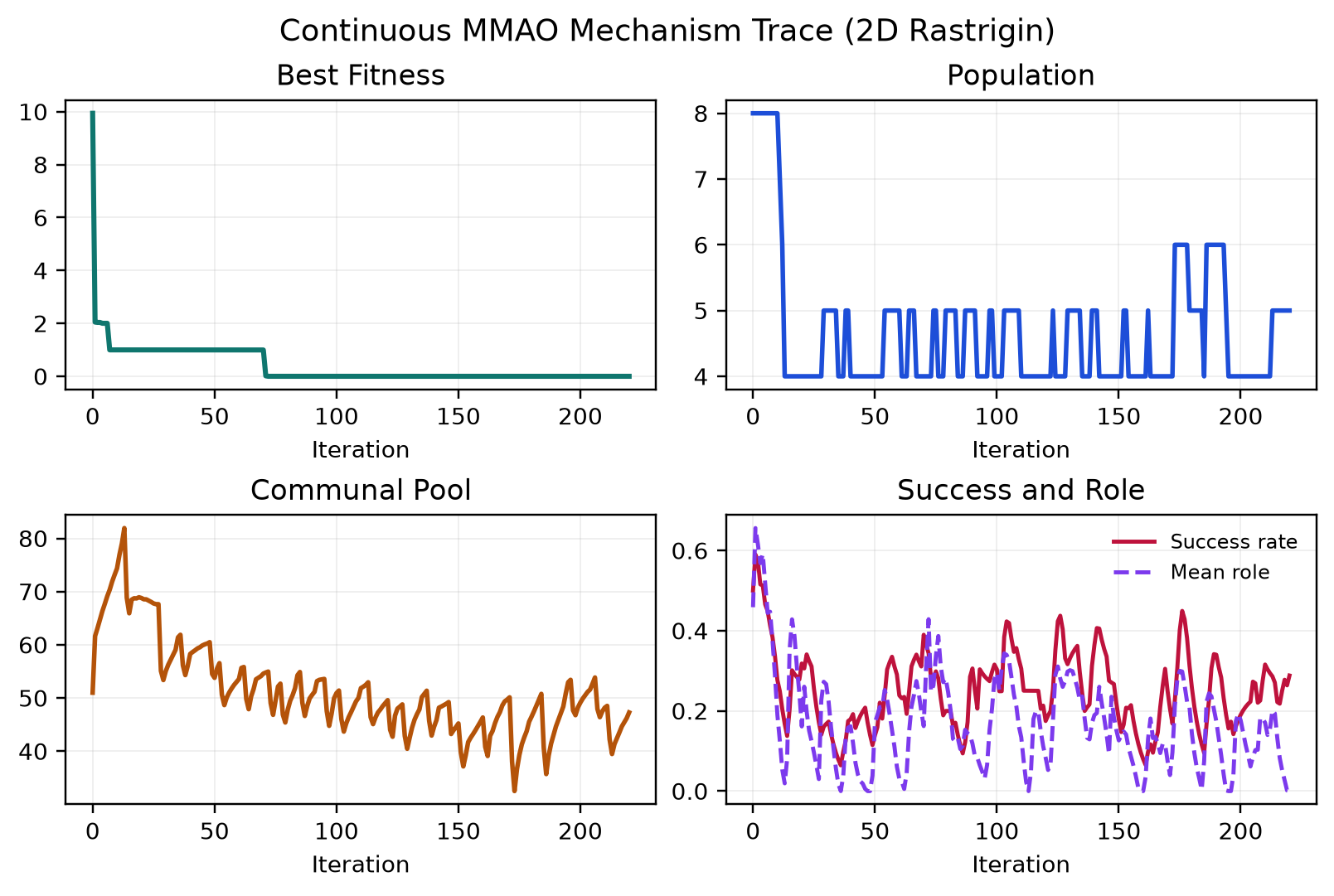}
\hfill
\includegraphics[width=0.48\textwidth]{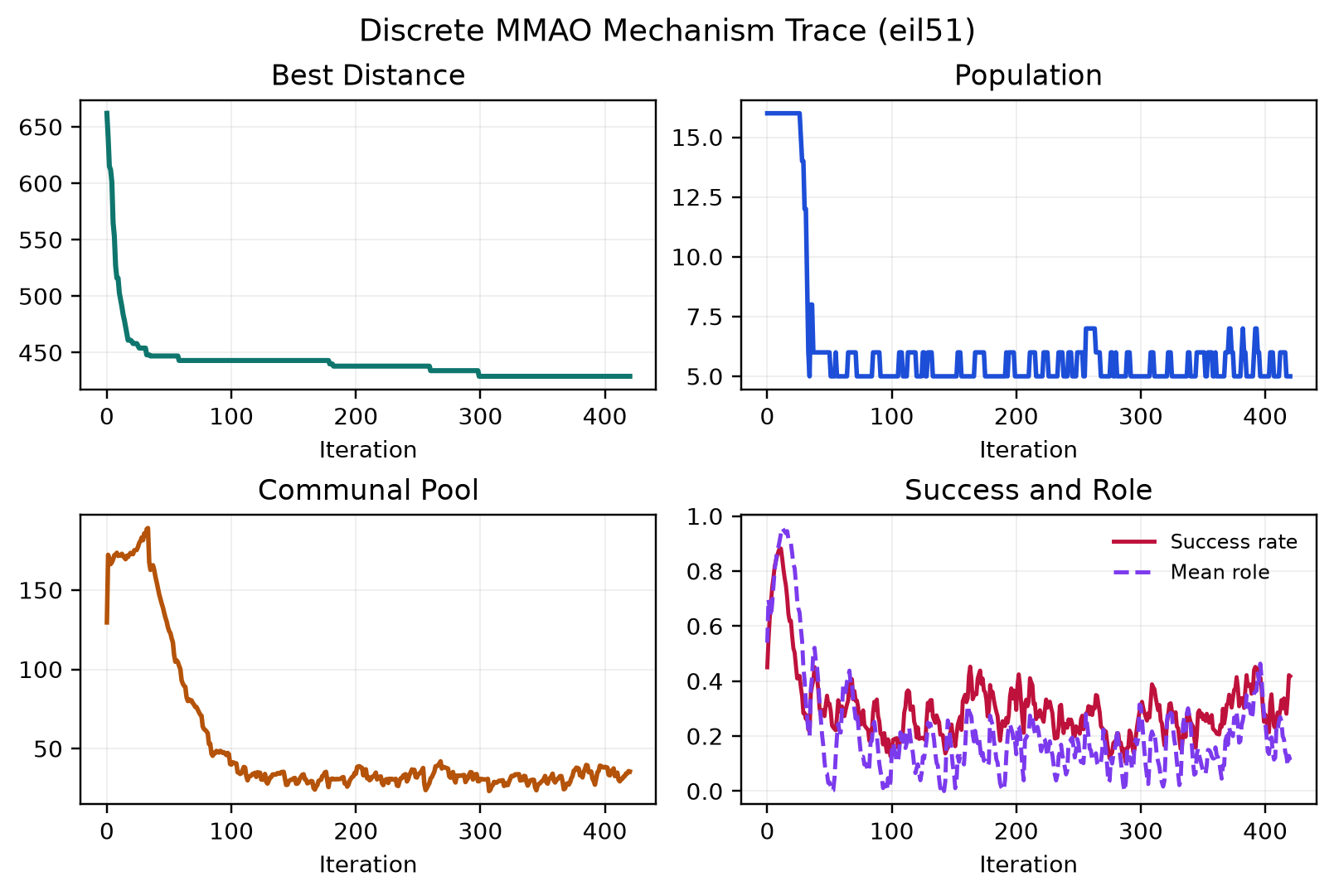}
\caption{Mechanism-level traces from two demonstration runs used to validate the closed-loop interpretation. (a) Continuous 2D Rastrigin: after rapid initial improvement, the active population contracts while the communal pool remains available for later reinvestment. (b) Discrete \texttt{eil51}: tour-length improvement is accompanied by sustained communal-budget turnover and role/success feedback fluctuations.}
\label{fig:mechanism-traces}
\end{figure*}

On the \texttt{eil51} demonstration, the trace shows the same pattern in structural form: the best tour length drops rapidly in the early phase, the communal pool is consumed as local reconstruction intensifies, and mean role remains biased toward exploitation while still fluctuating with success feedback (Fig.~\ref{fig:mechanism-traces}b).

\subsection{Baseline Comparisons}
Table~\ref{tab:continuous-baselines} and Table~\ref{tab:tsp-baselines} compare MMAO against lightweight representative baselines under the same benchmark setup. In the continuous suite, MMAO is much better than Random Search, Hill Climb, and DE-lite; all three Mann-Whitney U tests give extremely small $p$-values ($<10^{-20}$). In the discrete suite, MMAO also improves clearly over NN+2opt, RR-2opt, and CI+2opt on the aggregated benchmark set; the corresponding Mann-Whitney U tests yield $p=2.52\times10^{-34}$, $2.52\times10^{-34}$, and $5.41\times10^{-21}$, respectively.

\begin{table}[h]
\caption{Continuous baseline comparison averaged over the paper suite (320 runs per method). Lower is better.}
\label{tab:continuous-baselines}
\centering
\scriptsize
\begin{tabular}{lc}
\hline
\textbf{Method} & \textbf{Mean best-bias} \\
\hline
Random Search & $5.76 \times 10^{9}$ \\
Hill Climb & $7.19 \times 10^{9}$ \\
DE-lite & $7.34 \times 10^{8}$ \\
MMAO & $7.68 \times 10^{3}$ \\
\hline
\end{tabular}
\end{table}

\begin{table}[h]
\caption{TSPLIB baseline comparison averaged over the paper suite (100 runs per method across 5 instances). Lower gap is better.}
\label{tab:tsp-baselines}
\centering
\scriptsize
\begin{tabular}{lc}
\hline
\textbf{Method} & \textbf{Mean gap (\%)} \\
\hline
RR-2opt & $407.89$ \\
NN+2opt & $87.94$ \\
CI+2opt & $14.38$ \\
MMAO & $8.81$ \\
\hline
\end{tabular}
\end{table}

\begin{table}[h]
\caption{Rank-based baseline summary used to reduce cross-task scale bias. Lower rank is better.}
\label{tab:rank-summary}
\centering
\scriptsize
\begin{tabular}{lccc}
\hline
\textbf{Domain} & \textbf{Method} & \textbf{Avg. Rank} & \textbf{Wins} \\
\hline
Continuous & MMAO & 1.000 & 16/16 \\
Continuous & DE-lite & 2.125 & 0/16 \\
Continuous & Random Search & 3.188 & 0/16 \\
Continuous & Hill Climb & 3.688 & 0/16 \\
Discrete & MMAO & 1.000 & 5/5 \\
Discrete & CI+2opt & 2.000 & 0/5 \\
Discrete & NN+2opt & 3.000 & 0/5 \\
Discrete & RR-2opt & 4.000 & 0/5 \\
\hline
\end{tabular}
\end{table}

\begin{table}[h]
\caption{Formal statistical summary for the main comparisons.}
\label{tab:stat-summary}
\centering
\scriptsize
\begin{tabular}{lccc}
\hline
\textbf{Comparison} & \textbf{$p$-value} & \textbf{Cliff's $\delta$} & \textbf{CL} \\
\hline
Cont. MMAO vs Random Search & $2.60{\times}10^{-39}$ & -- & -- \\
Cont. MMAO vs Hill Climb & $1.92{\times}10^{-44}$ & -- & -- \\
Cont. MMAO vs DE-lite & $4.52{\times}10^{-21}$ & -- & -- \\
Disc. MMAO vs NN+2opt & $2.52{\times}10^{-34}$ & -- & -- \\
Disc. MMAO vs RR-2opt & $2.52{\times}10^{-34}$ & -- & -- \\
Disc. MMAO vs CI+2opt & $5.41{\times}10^{-21}$ & -- & -- \\
Cont. MMAO vs NoRoleDrift & $0.9261$ & $-0.0042$ & $0.5021$ \\
Cont. MMAO vs NoEliteReinvest & $0.1238$ & $-0.0703$ & $0.5351$ \\
Cont. MMAO vs WeakSuccessFeedback & $0.9461$ & $-0.0031$ & $0.5015$ \\
Disc. MMAO vs NoEdgeMemory & $0.3135$ & $-0.0825$ & $0.5413$ \\
Disc. MMAO vs NoGuidedReinvest & $0.6164$ & $-0.0410$ & $0.5205$ \\
Disc. MMAO vs WeakSuccessFeedback & $0.3266$ & $-0.0803$ & $0.5402$ \\
\hline
\end{tabular}
\end{table}

\subsection{Ablation Results}
The ablation study reveals which parts of the metabolic loop are essential and which parts are more problem-sensitive.

\begin{table}[h]
\caption{Continuous ablation summary averaged over the paper suite (320 runs per variant). Lower is better.}
\label{tab:continuous-ablation}
\centering
\scriptsize
\begin{tabular}{lc}
\hline
\textbf{Variant} & \textbf{Mean best-bias} \\
\hline
MMAO & $7.68 \times 10^{3}$ \\
NoRoleDrift & $1.04 \times 10^{4}$ \\
WeakSuccessFeedback & $7.28 \times 10^{3}$ \\
NoEliteReinvest & $2.56 \times 10^{4}$ \\
\hline
\end{tabular}
\end{table}

\begin{table}[h]
\caption{TSPLIB ablation summary averaged over the paper suite (100 runs per variant across 5 instances). Lower gap is better.}
\label{tab:tsp-ablation}
\centering
\scriptsize
\begin{tabular}{lc}
\hline
\textbf{Variant} & \textbf{Mean gap (\%)} \\
\hline
MMAO & $8.81$ \\
NoGuidedReinvest & $9.08$ \\
NoEdgeMemory & $9.59$ \\
WeakSuccessFeedback & $9.37$ \\
\hline
\end{tabular}
\end{table}

On the continuous side, the aggregate ablation picture remains mixed: removing elite reinvestment clearly hurts the mean result, whereas role drift and success-feedback weakening are more problem-sensitive in aggregate. On the discrete side, all three ablations are slightly worse than MMAO on average, but the effect sizes remain small: against MMAO, Cliff's delta is $-0.0825$ for NoEdgeMemory, $-0.0410$ for NoGuidedReinvest, and $-0.0803$ for WeakSuccessFeedback, with common-language probabilities around $0.52$--$0.54$ (Table~\ref{tab:stat-summary}). This is consistent with a controller whose components interact through one budget loop rather than acting as isolated tricks with uniformly additive gains.

\subsection{What the Current Experiments Establish}
The experiments support four conclusions. First, the MMAO implementation runs stably on a non-trivial continuous suite and on five standard TSPLIB instances. Second, the metabolic controller is useful relative to lightweight baselines: it is far better than simple generic baselines in the continuous suite and clearly stronger than NN+2opt, RR-2opt, and CI+2opt in the discrete suite. Third, the rank-based view in Table~\ref{tab:rank-summary} shows that this conclusion is not an artifact of averaging heterogeneous objective scales: MMAO ranks first on all 16 continuous tasks and all 5 discrete instances under the current protocol. Fourth, the paper should remain conservative and treat closed-loop resource allocation as the main validated idea; the present experiments do not justify claiming that every adaptive branch is already mature or universally beneficial.

The experiments also distinguish \emph{framework-level} evidence from \emph{instantiation-level} evidence. What is supported at the framework level is that one endogenous resource loop can regulate sensing intensity, role adaptation, reinvestment, pruning, and respawn across both Euclidean and permutation search spaces. What remains instantiation specific is how well a given move generator, refinement operator, or memory structure converts that common budget logic into final solution quality. This distinction matters because it explains why the present paper can establish MMAO as a credible optimizer without claiming that every current continuous or discrete operator is already the strongest possible realization of the framework.

\subsection{Reproducibility Workflow}
All results in this section are generated from the same reproducible experimental pipeline, with fixed seeds, unified benchmark definitions, and automatic regeneration of summary tables and diagnostic figures.

\section{Discussion}
The empirical results inform the interpretation of MMAO's resource-allocation mechanism. The key question is whether the current algorithm version succeeds in expressing a distinct and useful optimization principle: namely, that heterogeneous search behavior should be derived from a metabolically closed resource economy rather than from externally attached parameter controllers.

\subsection{Thermodynamics of Search: Energy as a Regularizer}
The most significant distinction of MMAO is its departure from static population models. In algorithms like PSO or DE, the population size $N$ is a fixed hyperparameter. This rigidity often leads to resource waste: in flat landscapes, redundant agents consume evaluations without progress; in complex multimodal landscapes, the fixed population may be insufficient to cover all attraction basins.

MMAO operates as a thermodynamic system where energy serves as a dynamic regularization term. In the revised formulation, the regularizer is not only the private maintenance cost but the entire private-public resource loop.
\begin{itemize}
    \item \textbf{Implicit Pruning:} Agents in unproductive regions eventually experience negative net balance and are removed, recycling part of their remaining energy into the communal pool.
    \item \textbf{Shared Reinvestment:} Productive regions do not only preserve successful agents; they also attract future sensing, branching, and respawn budget through the communal pool.
    \item \textbf{Self-regulated turnover:} The population expands during repeated success and contracts during stagnation, allowing the algorithm to adapt its search granularity without a handcrafted schedule \cite{signorelli2025perturbation}.
\end{itemize}

\subsection{From Optimization to Agent-Centered Search}
Traditional meta-heuristics treat individuals as passive particles governed by global formulas. In contrast, MMAO is closer to an agent-centered search view with a local perception-action loop \cite{pedreschi2025human, leuzzi2025lifelong, chen2024multi, mai2025constructive, sendra2025evolution, han2025robust, zhang2024virtual}. Each agent possesses a ``Perception-Action Loop'':
\begin{enumerate}
    \item \textbf{Perception:} The agent senses its local landscape under a budget determined by its current metabolic state.
    \item \textbf{Decision:} The role state $\phi_i$ converts success feedback and resource availability into a continuous exploratory-exploitative posture.
    \item \textbf{Action:} The motion or structural modification is generated from this role-conditioned budget.
\end{enumerate}
This agent-based autonomy enhances robustness against deceptive landscapes. Exploratory and exploitative behaviors coexist as metabolically regulated states inside one unified controller.

\subsection{Advantage of Symmetric Gradient Estimation}
The symmetric probing mechanism remains important in the continuous setting. By evaluating paired perturbations around the current point, MMAO obtains a more stable directional signal than purely one-sided random walk strategies \cite{nesterov2017random, seung2025low, hikima2025zeroth, lei2025zeroth}. The main methodological point, however, is not symmetric probing alone. Its scale, multiplicity, and downstream influence are all metabolically regulated by current spread, recent success, and available budget. The probing mechanism thus operates within the closed-loop controller.

\subsection{Why MMAO Is Parameter-Light Rather Than Naively Parameter-Free}
The present version of MMAO is best described as \textit{parameter-light}. It still contains implementation constants such as clipping ranges, smoothing rates, and minimum budgets. However, its most problem-sensitive behaviors---population resizing, search scale, structural budget, exploration pressure, role drift, and reinvestment bias---are not prescribed by handcrafted schedules. They are derived from the evolving metabolic state. This distinction is important and consistent with the broader parameter-control literature \cite{eiben1999parameter, karafotias2015parameter, brest2006self}. A meaningful parameter-free ambition for MMAO should mean that the algorithm does not require problem-specific manual tuning of search radii, population schedules, or explicit role thresholds, not that it contains literally no constants.

MMAO thus aims to minimize \emph{externally chosen behavioral schedules}. A user still inherits a compact set of implementation constants, but does not have to handcraft separate rules for when to expand the population, when to contract it, how much search effort to donate, when to intensify, or how aggressively to respawn. Those behaviors are coupled through the same metabolic accounting system. This is the practical sense in which MMAO is lighter than many adaptive meta-heuristics even before it becomes literally parameter free.

\subsection{Limitations and Threats to Validity}
MMAO in its current form still has several limitations and evidence-scope constraints.
\begin{itemize}
    \item \textbf{Benchmark scope:} The current paper covers a meaningful but still limited benchmark slice: eight CEC2017 functions at 10D/30D and five TSPLIB instances. This is enough to validate workability, but not enough to claim broad superiority across the full meta-heuristic landscape.
    \item \textbf{Budget comparability:} The comparisons are protocol-matched rather than exactly function-evaluation matched across all optimizer families. This is acceptable for the present proof-of-concept study, but tighter cross-family budget normalization would strengthen future comparisons.
    \item \textbf{Discrete ambition:} The TSP instantiation is intentionally generic and should not be expected to match highly specialized solvers such as LKH without further structural innovation.
    \item \textbf{Mechanism maturity:} The new ablation results show that not every adaptive branch is uniformly beneficial. In particular, role drift and guided reinvestment remain problem-sensitive on the current suite, so future work should focus on deriving a smaller set of stronger endogenous rules rather than adding more controller layers.
    \item \textbf{Theory and constants:} The present theory explains boundedness, normalization, and turnover but does not yet provide a tight finite-time characterization of the full adaptive system. In addition, some implementation constants still reflect engineering judgment rather than first-principles derivation, so the parameter-free ambition is only partially realized.
\end{itemize}

These limitations also clarify what the present paper is and is not trying to complete. It is intended to define the founding MMAO controller and to show that the same closed-loop logic can be instantiated credibly in both continuous and discrete optimization. It is not yet a large-scale validation paper, a dynamic-optimization paper, or a mixed-space application paper. Those directions are natural consequences of the framework if they preserve the same resource language, but they should be treated as distinct research questions rather than silently folded into the claim scope of the present work.

\section{Conclusion}
In this paper, we reconstructed the Metabolic Multi-Agent Optimizer (MMAO) around its current algorithmic core: a private-public metabolic resource loop that governs sensing, role adaptation, action generation, branching, pruning, respawn, and reinvestment. The formulation specifies bounded normalized reward, continuous heterogeneous behavior, and unified resource allocation.

MMAO regulates heterogeneous optimization behavior in both continuous and discrete domains through the same endogenous resource economy. The current evidence supports this claim on the tested CEC2017 subset and on five TSPLIB instances, and the strongest conclusion is therefore that the closed-loop resource-allocation idea is workable and useful. This makes MMAO a promising basis for further work on parameter-light, self-calibrating optimization, while still leaving ample room for simplification, stronger theory, and stronger domain-specific validation.

The formulation also provides a reference for subsequent MMAO research by specifying the framework's metabolic bookkeeping requirements. That clarification is valuable even where later studies revise operators, strengthen evidence, or simplify the mechanism set.

Future work should focus on stronger large-scale benchmarking, tighter theory for the fully adaptive system, and broader application domains such as dynamic optimization and neural architecture search \cite{wu2024evolutionary}. Further adaptive behaviors should be derived from the metabolic loop.

\section*{Acknowledgement}
The related publications in Table~\ref{tab:mmao-research-series} provide complementary views of the MMAO research series.

\begin{table}[!ht]
\centering
\caption{Other publications in the MMAO research series.}
\label{tab:mmao-research-series}
\footnotesize
\setlength{\tabcolsep}{5pt}
\renewcommand{\arraystretch}{1.12}
\begin{tabular}{@{}p{\columnwidth}@{}}
\hline
\textbf{Related publication} \\
\hline
\emph{Minimal MMAO: A Resource-Closed-Loop Framework for Adaptive Metaheuristic Search}~\cite{xu2026minimalmmao}.\newline
\href{https://arxiv.org/abs/2606.30450}{arXiv:2606.30450}
\\[5pt]
\emph{A Large-Scale Empirical Evaluation of MMAO Under Fair-Budget Continuous and Discrete Benchmarks}~\cite{xu2026largescale}.\newline
\href{https://arxiv.org/abs/2606.31584}{arXiv:2606.31584}
\\[5pt]
\emph{MMAO-Dyn: A Metabolic Multi-Agent Optimizer for Dynamic Optimization}~\cite{xu2026mmaodyn}.\newline
\href{https://arxiv.org/abs/2607.00846}{arXiv:2607.00846}
\\[5pt]
\emph{MMAO-Cls: Metabolic Multi-Agent Optimization for Joint Feature Selection and Classifier Tuning}~\cite{xu2026mmaocls}.\newline
\href{https://arxiv.org/abs/2607.01539}{arXiv:2607.01539}
\\[5pt]
\emph{Mechanism and Stability Analysis of Metabolic Closed-Loop Metaheuristics}~\cite{xu2026mechanism}.\newline
\href{https://arxiv.org/abs/2607.01551}{arXiv:2607.01551}
\\[5pt]
\emph{MMAO: A Unified Resource-Regulated Controller for Cross-Domain Optimization}~\cite{xu2026unifiedmmao}.\newline
\href{https://doi.org/10.1109/ACCESS.2026.3752083}{IEEE Access, 2026}
\\[5pt]
\hline
\end{tabular}
\end{table}

\bibliographystyle{ieeetr}
\bibliography{Ref}

\begin{thebibliography}{10}

\bibitem{jiang2023knowledge}
Y.~Jiang, Z.-H. Zhan, K.~C. Tan, and J.~Zhang, ``Knowledge learning for
  evolutionary computation,'' {\em IEEE transactions on evolutionary
  computation}, vol.~29, no.~1, pp.~16--30, 2023.

\bibitem{shao2025knowledge}
S.~Shao, Y.~Tian, Y.~Zhang, and X.~Zhang, ``Knowledge learning-based
  dimensionality reduction for solving large-scale sparse multiobjective
  optimization problems,'' {\em IEEE Transactions on Cybernetics}, 2025.

\bibitem{kennedy1995particle}
J.~Kennedy and R.~Eberhart, ``Particle swarm optimization,'' in {\em
  Proceedings of ICNN'95-international conference on neural networks}, vol.~4,
  pp.~1942--1948, ieee, 1995.

\bibitem{doerr2025speeding}
B.~Doerr, M.~S. Krejca, and S.~Wietheger, ``Speeding up the nsga-ii via dynamic
  population sizes,'' in {\em Proceedings of the AAAI Conference on Artificial
  Intelligence}, pp.~26964--26972, 2025.

\bibitem{antipov2024already}
D.~Antipov, B.~Doerr, and A.~Ivanova, ``Already moderate population sizes
  provably yield strong robustness to noise,'' in {\em Proceedings of the
  Genetic and Evolutionary Computation Conference}, pp.~1524--1532, 2024.

\bibitem{li2025scalable}
M.~Li, W.~Zheng, and B.~Doerr, ``Scalable speed-ups for the sms-emoa from a
  simple aging strategy,'' in {\em Proceedings of the Thirty-Fourth
  International Joint Conference on Artificial Intelligence, {IJCAI-25}}, 2025.

\bibitem{bian2025archive}
C.~Bian, S.~Ren, M.~Li, and C.~Qian, ``An archive can bring provable speed-ups
  in multi-objective evolutionary algorithms,'' in {\em Proceedings of the
  Genetic and Evolutionary Computation Conference Companion}, pp.~6905--6913,
  2024.

\bibitem{bian2025stochastic}
C.~Bian, Y.~Zhou, M.~Li, and C.~Qian, ``Stochastic population update can
  provably be helpful in multi-objective evolutionary algorithms,'' {\em
  Artificial Intelligence}, vol.~341, p.~104308, 2025.

\bibitem{Tanabe2014}
R.~Tanabe and A.~S. Fukunaga, ``Improving the search performance of shade using
  linear population size reduction,'' in {\em 2014 IEEE Congress on
  Evolutionary Computation (CEC)}, pp.~1658--1665, IEEE, 2014.

\bibitem{Hansen2001}
N.~Hansen and A.~Ostermeier, ``Completely derandomized self-adaptation in
  evolution strategies,'' {\em Evolutionary Computation}, vol.~9, no.~2,
  pp.~159--195, 2001.

\bibitem{cho2025configx}
H.~Guo, Z.~Ma, J.~Chen, Y.~Ma, Z.~Cao, X.~Zhang, and Y.-J. Gong, ``Configx:
  Modular configuration for evolutionary algorithms via multitask reinforcement
  learning,'' in {\em Proceedings of the AAAI Conference on Artificial
  Intelligence}, vol.~39, pp.~26982--26990, 2025.

\bibitem{zhang2025laos}
Y.~Zhang and G.~Yi, ``Laos: Large language model-driven adaptive operator
  selection for evolutionary algorithms,'' in {\em Proceedings of the Genetic
  and Evolutionary Computation Conference}, pp.~517--526, 2025.

\bibitem{song2024reinforced}
L.~Song, C.~Gao, K.~Xue, C.~Wu, D.~Li, J.~Hao, Z.~Zhang, and C.~Qian,
  ``Reinforced in-context black-box optimization,'' in {\em Proceedings of the
  Thirty-Third International Joint Conference on Artificial Intelligence,
  {IJCAI-24}}, 2024.

\bibitem{eiben1999parameter}
A.~E. Eiben, R.~Hinterding, and Z.~Michalewicz, ``Parameter control in
  evolutionary algorithms,'' {\em IEEE Transactions on Evolutionary
  Computation}, vol.~3, no.~2, pp.~124--141, 1999.

\bibitem{karafotias2015parameter}
G.~Karafotias, M.~Hoogendoorn, and A.~E. Eiben, ``Parameter control in
  evolutionary algorithms: Trends and challenges,'' {\em IEEE Transactions on
  Evolutionary Computation}, vol.~19, no.~2, pp.~167--187, 2015.

\bibitem{dong2025effective}
Z.~Dong and X.~Wang, ``Effective computational resource allocation in
  evolutionary multi-objective multi-task optimization,'' in {\em 2025 IEEE
  Congress on Evolutionary Computation (CEC)}, pp.~1--7, IEEE, 2025.

\bibitem{li2022distributed}
J.-Y. Li, K.-J. Du, Z.-H. Zhan, H.~Wang, and J.~Zhang, ``Distributed
  differential evolution with adaptive resource allocation,'' {\em IEEE
  transactions on cybernetics}, vol.~53, no.~5, pp.~2791--2804, 2022.

\bibitem{liu2022cooperative}
X.-F. Liu, J.~Zhang, and J.~Wang, ``Cooperative particle swarm optimization
  with a bilevel resource allocation mechanism for large-scale dynamic
  optimization,'' {\em IEEE Transactions on Cybernetics}, vol.~53, no.~2,
  pp.~1000--1011, 2022.

\bibitem{chu2024competitive}
X.~Chu, F.~Ming, and W.~Gong, ``Competitive multitasking for computational
  resource allocation in evolutionary constrained multi-objective
  optimization,'' {\em IEEE Transactions on Evolutionary Computation}, 2024.

\bibitem{han2022multitask}
H.~Han, X.~Bai, Y.~Hou, and J.~Qiao, ``Multitask particle swarm optimization
  with dynamic on-demand allocation,'' {\em IEEE Transactions on Evolutionary
  Computation}, vol.~27, no.~4, pp.~1015--1026, 2022.

\bibitem{elorza2024transforming}
A.~Elorza, X.~Benavides, J.~Ceberio, L.~Hernando, and J.~A. Lozano,
  ``Transforming combinatorial optimization problems in fourier space:
  Consequences and uses,'' {\em IEEE Transactions on Evolutionary Computation},
  vol.~29, no.~4, pp.~977--989, 2024.

\bibitem{liu2025less}
S.~Liu, Z.-J. Wang, Z.~Kou, Z.-H. Zhan, S.~Kwong, and J.~Zhang, ``Less is more:
  A small-scale learning particle swarm optimization for large-scale
  optimization,'' {\em IEEE Transactions on Cybernetics}, vol.~56, no.~1,
  pp.~523--536, 2026.

\bibitem{mai2025constructive}
Y.~Mai, Q.~Yin, W.~Ni, P.~Xu, and K.~Huang, ``Constructive conflict-driven
  multi-agent reinforcement learning for strategic diversity,'' in {\em
  Proceedings of the Thirty-Fourth International Joint Conference on Artificial
  Intelligence, {IJCAI-25}}, 2025.

\bibitem{sendra2025evolution}
R.~Sendra-Arranz, {\'A}.~Guti{\'e}rrez, and A.~L. Christensen, ``Evolution of
  transferable and self-organized communication modules for solving multiple
  swarm robotics tasks,'' {\em IEEE Transactions on Cybernetics}, 2025.

\bibitem{han2025robust}
H.~Han, H.~Zhou, Y.~Huang, and Y.~Hou, ``Robust multiobjective competitive
  swarm optimization based on evolutionary trend prediction,'' {\em IEEE
  Transactions on Cybernetics}, 2025.

\bibitem{chen2024multi}
T.-Y. Chen, W.-N. Chen, F.-F. Wei, X.-M. Hu, and J.~Zhang, ``Multi-agent swarm
  optimization with adaptive internal and external learning for complex
  consensus-based distributed optimization,'' {\em IEEE Transactions on
  Evolutionary Computation}, 2024.

\bibitem{zhang2024virtual}
J.~Zhang, Y.~Lin, and M.~Zhou, ``Virtual-source and virtual-swarm-based
  particle swarm optimizer for large-scale multi-source location via robot
  swarm,'' {\em IEEE Transactions on Evolutionary Computation}, 2024.

\bibitem{pedreschi2025human}
D.~Pedreschi, L.~Pappalardo, E.~Ferragina, R.~Baeza-Yates, A.-L. Barab{\'a}si,
  F.~Dignum, V.~Dignum, T.~Eliassi-Rad, F.~Giannotti, J.~Kert{\'e}sz, {\em
  et~al.}, ``Human-ai coevolution,'' {\em Artificial Intelligence}, vol.~339,
  p.~104244, 2025.

\bibitem{leuzzi2025lifelong}
L.~Leuzzi, D.~Bacciu, S.~Hauert, S.~Jones, and A.~Cossu, ``Lifelong evolution
  of swarms,'' in {\em Proceedings of the Genetic and Evolutionary Computation
  Conference}, pp.~1549--1557, 2025.

\bibitem{wang2023multiobjective}
Z.~Wang, S.~Yao, G.~Li, and Q.~Zhang, ``Multiobjective combinatorial
  optimization using a single deep reinforcement learning model,'' {\em IEEE
  transactions on cybernetics}, vol.~54, no.~3, pp.~1984--1996, 2023.

\bibitem{xu2023cumulative}
X.~Xu, J.~Cao, X.~Shi, and S.~Gorbachev, ``Cumulative capacitated colored
  traveling salesman problem,'' {\em IEEE Transactions on Cybernetics},
  vol.~54, no.~8, pp.~4553--4566, 2023.

\bibitem{fu2023hierarchical}
J.~Fu, G.~Sun, J.~Liu, W.~Yao, and L.~Wu, ``On hierarchical multi-uav dubins
  traveling salesman problem paths in a complex obstacle environment,'' {\em
  IEEE Transactions on Cybernetics}, vol.~54, no.~1, pp.~123--135, 2023.

\bibitem{zhao2024pega}
B.~Zhao, W.-N. Chen, F.-F. Wei, X.~Liu, Q.~Pei, and J.~Zhang, ``Pega: A
  privacy-preserving genetic algorithm for combinatorial optimization,'' {\em
  IEEE Transactions on Cybernetics}, vol.~54, no.~6, pp.~3638--3651, 2024.

\bibitem{verdu2025scaling}
F.~J.~C. Verd{\`u}, L.~Castelli, and L.~Bortolussi, ``Scaling combinatorial
  optimization neural improvement heuristics with online search and
  adaptation,'' in {\em Proceedings of the AAAI Conference on Artificial
  Intelligence}, vol.~39, pp.~27135--27143, 2025.

\bibitem{ye2023deepaco}
H.~Ye, J.~Wang, Z.~Cao, H.~Liang, and Y.~Li, ``Deepaco: Neural-enhanced ant
  systems for combinatorial optimization,'' in {\em Advances in Neural
  Information Processing Systems}, vol.~36, 2023.

\bibitem{doerr2023understanding}
B.~Doerr and Z.~Qu, ``From understanding the population dynamics of the nsga-ii
  to the first proven lower bounds,'' in {\em Proceedings of the AAAI
  Conference on Artificial Intelligence}, vol.~37, pp.~12408--12416, 2023.

\bibitem{dang2025dominance}
D.-C. Dang, A.~Opris, and D.~Sudholt, ``Why dominance is not enough: Lessons
  from practical evolutionary multi-objective algorithms,'' in {\em Proceedings
  of the Genetic and Evolutionary Computation Conference}, pp.~1604--1612,
  2025.

\bibitem{zheng2023runtime}
W.~Zheng and B.~Doerr, ``Runtime analysis for the nsga-ii: proving,
  quantifying, and explaining the inefficiency for many objectives,'' {\em IEEE
  Transactions on Evolutionary Computation}, vol.~28, no.~5, pp.~1442--1454,
  2023.

\bibitem{vermetten2024large}
D.~Vermetten, C.~Doerr, H.~Wang, A.~V. Kononova, and T.~B{\"a}ck, ``Large-scale
  benchmarking of metaphor-based optimization heuristics,'' in {\em proceedings
  of the genetic and evolutionary computation conference}, pp.~41--49, 2024.

\bibitem{cenikj2025comparing}
G.~Cenikj, G.~Petelin, and T.~Eftimov, ``Comparing optimization algorithms
  through the lens of search behavior analysis,'' in {\em Proceedings of the
  Genetic and Evolutionary Computation Conference Companion}, pp.~475--478,
  2025.

\bibitem{lehre2024concentration}
P.~K. Lehre and S.~Lin, ``Concentration tail-bound analysis of coevolutionary
  and bandit learning algorithms,'' in {\em Proceedings of the Thirty-Third
  International Joint Conference on Artificial Intelligence, {IJCAI-24}}, 2024.

\bibitem{omeradzic2024self}
A.~Omeradzic and H.-G. Beyer, ``Self-adaptation of multi-recombinant evolution
  strategies on the highly multimodal rastrigin function,'' {\em IEEE
  Transactions on Evolutionary Computation}, 2024.

\bibitem{prager2024exploratory}
R.~P. Prager and H.~Trautmann, ``Exploratory landscape analysis for
  mixed-variable problems,'' {\em IEEE Transactions on Evolutionary
  Computation}, 2024.

\bibitem{stripinis2024benchmarking}
L.~Stripinis, J.~K{\r{u}}dela, and R.~Paulavi{\v{c}}ius, ``Benchmarking
  derivative-free global optimization algorithms under limited dimensions and
  large evaluation budgets,'' {\em IEEE Transactions on Evolutionary
  Computation}, vol.~29, no.~1, pp.~187--204, 2024.

\bibitem{raponi2023optimizing}
E.~Raponi, N.~C. Rakotonirina, J.~Rapin, C.~Doerr, and O.~Teytaud, ``Optimizing
  with low budgets: A comparison on the black-box optimization benchmarking
  suite and openai gym,'' {\em IEEE Transactions on Evolutionary Computation},
  vol.~29, no.~1, pp.~91--101, 2023.

\bibitem{zheng2024approximation}
W.~Zheng and B.~Doerr, ``Approximation guarantees for the non-dominated sorting
  genetic algorithm ii (nsga-ii),'' {\em IEEE Transactions on Evolutionary
  Computation}, 2024.

\bibitem{holland1975adaptation}
J.~H. Holland, {\em Adaptation in Natural and Artificial Systems}.
\newblock Ann Arbor, MI: The University of Michigan Press, 1975.

\bibitem{kirkpatrick1983optimization}
S.~Kirkpatrick, C.~D. Gelatt~Jr, and M.~P. Vecchi, ``Optimization by simulated
  annealing,'' {\em science}, vol.~220, no.~4598, pp.~671--680, 1983.

\bibitem{li2020slime}
S.~Li, H.~Chen, M.~Wang, A.~A. Heidari, and S.~Mirjalili, ``Slime mould
  algorithm: A new method for stochastic optimization,'' {\em Future generation
  computer systems}, vol.~111, pp.~300--323, 2020.

\bibitem{cui2013artificial}
Z.~Cui and X.~Cai, ``Artificial plant optimization algorithm,'' in {\em Swarm
  Intelligence and Bio-Inspired Computation}, pp.~351--365, Elsevier, 2013.

\bibitem{kiran2015tsa}
M.~S. Kiran, ``Tsa: Tree-seed algorithm for continuous optimization,'' {\em
  Expert Systems with Applications}, vol.~42, no.~19, pp.~6686--6698, 2015.

\bibitem{azizipour2016optimal}
M.~Azizipour, V.~Ghalenoei, M.~Afshar, and S.~Solis, ``Optimal operation of
  hydropower reservoir systems using weed optimization algorithm,'' {\em Water
  resources management}, vol.~30, no.~11, pp.~3995--4009, 2016.

\bibitem{qi2013idea}
X.~Qi, Y.~Zhu, H.~Chen, D.~Zhang, and B.~Niu, ``An idea based on plant root
  growth for numerical optimization,'' in {\em International Conference on
  Intelligent Computing}, pp.~571--578, Springer, 2013.

\bibitem{zhang2012root}
H.~Zhang, Y.~Zhu, and H.~Chen, ``Root growth model for simulation of plant root
  system and numerical function optimization,'' in {\em International
  Conference on Intelligent Computing}, pp.~641--648, Springer, 2012.

\bibitem{nesterov2017random}
Y.~Nesterov and V.~Spokoiny, ``Random gradient-free minimization of convex
  functions,'' {\em Foundations of Computational Mathematics}, vol.~17, no.~2,
  pp.~527--566, 2017.

\bibitem{seung2025low}
H.~Seung, J.~Lee, and H.~Ko, ``Low-rank curvature for zeroth-order optimization
  in llm fine-tuning,'' in {\em Proceedings of the AAAI Conference on
  Artificial Intelligence}, pp.~1--10, AAAI Press, 2025.

\bibitem{hikima2025zeroth}
Y.~Hikima and A.~Takeda, ``Zeroth-order methods for nonconvex stochastic
  problems with decision-dependent distributions,'' in {\em Proceedings of the
  AAAI Conference on Artificial Intelligence}, vol.~39, pp.~17195--17203, 2025.

\bibitem{lei2025zeroth}
Y.~Lei, Y.~Lyu, G.~Zhan, T.~Zhang, J.~Li, J.~Chen, S.~E. Li, and S.~Zheng,
  ``Zeroth-order actor--critic: An evolutionary framework for sequential
  decision problems,'' {\em IEEE Transactions on Evolutionary Computation},
  vol.~29, no.~2, pp.~555--569, 2025.

\bibitem{li2014adaptivebandit}
K.~Li, {\'A}.~Fialho, S.~Kwong, and Q.~Zhang, ``Adaptive operator selection
  with bandits for a multiobjective evolutionary algorithm based on
  decomposition,'' {\em IEEE Transactions on Evolutionary Computation},
  vol.~18, no.~1, pp.~114--130, 2014.

\bibitem{heins2025repair}
J.~Heins, D.~Whitley, and P.~Kerschke, ``To repair or not to repair?
  investigating the importance of ab-cycles for the state-of-the-art tsp
  heuristic eax,'' in {\em Proceedings of the Genetic and Evolutionary
  Computation Conference}, pp.~231--239, 2025.

\bibitem{wu2024reinforcens}
T.~Wu, H.~Cui, T.~Guan, Y.~Wang, and Y.~Jin, ``Reinforcens: Reinforcement
  learning-based multi-start neighborhood search for solving the traveling
  thief problem.,'' in {\em Proceedings of the Thirty-Third International Joint
  Conference on Artificial Intelligence, {IJCAI-24}}, pp.~7038--7046, 2024.

\bibitem{signorelli2025perturbation}
F.~Signorelli and A.~Yaman, ``A perturbation and speciation-based algorithm for
  dynamic optimization uninformed of change,'' in {\em Proceedings of the
  Genetic and Evolutionary Computation Conference}, pp.~773--781, 2025.

\bibitem{brest2006self}
J.~Brest, S.~Greiner, B.~Boskovic, M.~Mernik, and V.~Zumer, ``Self-adapting
  control parameters in differential evolution: A comparative study on
  numerical benchmark problems,'' {\em IEEE Transactions on Evolutionary
  Computation}, vol.~10, no.~6, pp.~646--657, 2006.

\bibitem{wu2024evolutionary}
X.~Wu, S.-h. Wu, J.~Wu, L.~Feng, and K.~C. Tan, ``Evolutionary computation in
  the era of large language model: Survey and roadmap,'' {\em IEEE Transactions
  on Evolutionary Computation}, vol.~29, no.~2, pp.~534--554, 2024.

\bibitem{xu2026minimalmmao}
J.~Xu and L.~Ma, ``Minimal mmao: A resource-closed-loop framework for adaptive
  metaheuristic search,'' 2026.

\bibitem{xu2026largescale}
J.~Xu and L.~Ma, ``A large-scale empirical evaluation of mmao under fair-budget
  continuous and discrete benchmarks,'' 2026.

\bibitem{xu2026mmaodyn}
J.~Xu and L.~Ma, ``Mmao-dyn: A metabolic multi-agent optimizer for dynamic
  optimization,'' 2026.

\bibitem{xu2026mmaocls}
J.~Xu and L.~Ma, ``Mmao-cls: Metabolic multi-agent optimization for joint
  feature selection and classifier tuning,'' 2026.

\bibitem{xu2026mechanism}
J.~Xu and L.~Ma, ``Mechanism and stability analysis of metabolic closed-loop
  metaheuristics,'' 2026.

\bibitem{xu2026unifiedmmao}
J.~Xu, B.~Li, L.~Ma, R.~Zhu, F.~Nie, and J.~Li, ``{MMAO}: A unified
  resource-regulated controller for cross-domain optimization,'' {\em IEEE
  Access}, pp.~1--12, 2026.

\end{thebibliography}

\end{document}